\pdfoutput=1

\documentclass[11pt]{article}

\usepackage[final]{acl}

\usepackage{times}
\usepackage{latexsym}
\usepackage{multirow}
\usepackage{makecell}
 \usepackage{enumitem}

\usepackage[T1]{fontenc}

\usepackage[utf8]{inputenc}

\usepackage{microtype}

\usepackage{inconsolata}

\usepackage{graphicx}
\usepackage{amsmath}
\graphicspath{ {./images/} }

\title{[Preprint] Non-Monotonic Attention-based Read/Write Policy Learning for Simultaneous Translation}

\author{Zeeshan Ahmed, Frank Seide, Zhe Liu, Rastislav Rabatin\thanks{work done while at Meta Inc.}, Jachym Kolar\thanks{work done while at Meta Inc.}, \\ {\bf Niko Moritz}, {\bf Ruiming Xie}, {\bf Simone Merello \and}{\bf Christian Fuegen} \\
        Meta AI}

\begin{document}
\maketitle
\begin{abstract}
Simultaneous or streaming machine translation generates translation while reading the input stream. These systems face a quality/latency trade-off, aiming to achieve high translation quality similar to non-streaming models with minimal latency. We propose an approach that efficiently manages this trade-off. By enhancing a pretrained non-streaming model, which was trained with a seq2seq mechanism and represents the upper bound in quality, we convert it into a streaming model by utilizing the alignment between source and target tokens. This alignment is used to learn a read/write decision boundary for reliable translation generation with minimal input. During training, the model learns the decision boundary through a read/write policy module, employing supervised learning on the alignment points (pseudo labels). The read/write policy module, a small binary classification unit, can control the quality/latency trade-off during inference. Experimental results show that our model outperforms several strong baselines and narrows the gap with the non-streaming baseline model.
\end{abstract}

\section{Introduction}

Simultaneous or streaming machine translation initiates the translation process while still receiving the input. This contrasts with conventional translation systems, which waits to process the entire input before starting the translation. Simultaneous translation is particularly beneficial for translating live speech in real-time, as well as in two-party conversations where users want to hear their partner promptly to have a more natural conversation even if they speak different languages.

Standard non-streaming machine translation systems rely on an encoder-decoder architecture, which typically requires processing the entire input before starting to translate. The decoder uses an attention module that sees all encoder outputs. For simultaneous translation, both the encoder and attention module must be adaptive, allowing the encoder to operate incrementally and the cross-attention in the decoder to handle variable-length encoder output sequences. In this setup, the streaming model is fed one token at a time, and at each decoding step, it decides whether it wants to wait for another token or write a token to the output. The simultaneous translation systems face a trade-off between quality and latency. Prioritizing higher quality increases latency (with non-streaming systems still representing the upper bound in quality), while reducing latency can impact translation quality. Therefore, simultaneous systems must carefully balance this trade-off for practical applications.

With the recent approaches \cite{emma2023, mma2019, milk2019, mocha2018, raffel2017online}, streaming machine translation is achieved with an alternative attention mechanism called Monotonic Attention (MA). The MA doesn't have access to the full sentence thus can only attend to the previous encoder states. The MA has an attention component together with a policy component. The attention component job is still to attend to the relevant pieces of the available input while the policy learner on the other hand learns a policy to decide whether to read more input or generate the translation (write). This read/write policy decision makes the model capable of generating a translation as soon as the model has seen enough information at the input. 

In addition to the MA approach, there are approaches which use a fixed size policy. For example, the wait-$k$ approach which waits for the $k$ input tokens to arrive before generating the translation. It then alternates between read/write operations based on the read/write probability \cite{dalvi2018incremental, ma2019stacl}. Even a non-streaming model has been used for streaming translation where the system uses some heuristics to estimate the threshold based on MT confidence to drive the read/write policy decision \cite{cho2016can}.

We present a novel approach to simultaneous translation called ALIgnment BAsed STReaming Machine Translation (AliBaStr-MT) which leverages a high quality pretrained non-streaming model as a starting point and uses the \textbf{non-monotonic attention as monotonic attention} when equipped with a learnable read/write module. The read/write module is a binary classifier which is trained based on the alignment obtained from the pretrained model.

These are the advantages of our approach:
\begin{itemize}[itemsep=0.5mm]
  \item \textbf{Training efficiency}: the model doesn't need to be retrained or trained with expensive operations as in MMA model. Only a light weight read/write classifier is trained with alignment information from the base non-streaming model (which is also used to mask the future encoder states during training).
  \item \textbf{Inference efficiency}: computation of read/write probability for MMA model during inference is expensive. In our case, we only need to run a small classifier on the top of the model which is quite fast.
  \item \textbf{Easy fallback to non-streaming model}: The base non-streaming model either remains unchanged or trained together with a read/write module. It offers the flexibility to still use the base model as the non-streaming model when high quality is required. 
  \item \textbf{Flexibility between latency and accuracy trade-off}: The calibration threshold of the read/write binary classifier becomes the inference time hyper-parameter instead of training time parameter as in monotonic attention i.e. changing the parameter value requires retraining the model. In our approach, it allows us to tune the model for the quality/latency trade-off during inference time.
\end{itemize}


\section{Background}
The simultaneous translation systems can be broadly classified into fixed policy and adaptive policy systems. In the fixed policy system, the model uses the predefined heuristic and rules to generate the translation. These heuristics remain constant throughout the translation. The fixed policy systems such as wait-$k$ or prefix-to-prefix translation systems by \cite{dalvi2018incremental, ma2019stacl} are simple and yet effective methods for streaming translation. However, the value for $k$ in wait-$k$ is not a learnable parameter of the model and set beforehand making it difficult to adjust based on the input and so far generated translation. The $k$ is also language dependent for different translation directions making it difficult to build a multilingual streaming translation model with wait-$k$. \cite{guo2024glancing} on the other hand tries to fix the quality gap in simultaneous translation models by introducing a look-ahead feature and the curriculum learning during training.

The adaptive policy methods  on the other hand dynamically adjust the read/write decision based on the input and previously generated translation. The adaptive method includes the ones where non-streaming MT model (trained on full sentences) is kept fixed/frozen and a heuristics or learnable module is added to make read/write decisions e.g. the confidence threshold (What-if-*) method of \cite{cho2016can} and reinforcement learning approaches by \cite{grissom2014don, gu2017learning} use pretrained non-streaming model which is not modified during the training. \cite{press2018you} also proposed an adaptive system without attention. The model keeps the single context vector to reduce computational and memory footprint. In addition to the target tokens, the model can also generate the empty symbol to delay the translation. However, the model requires external supervision in the form of alignment between the source and target words to train the system. 

Most of the recent work in adaptive policy methods is based on monotonic attention. The monotonic attention in essence learns a dynamic read/write policy which varies over the course of translation. There are various flavors available for monotonic attention e.g. Monotonic Hard Attention \cite{raffel2017online} where the model only attends to the latest encoder state causing improvement in the latency and runtime while making compromise on the translation quality. Monotonic Chunkwise Attention (MoChA) \cite{mocha2018} on the other hand attends to the segment of previous encoder states to overcome the quality gap. Monotonic Infinite Lookback (MILK) \cite{milk2019} further improves upon the quality by attending to the entire previous encoder states. These flavors can also be combined with their multi-head version called Monotonic Multi-head Attention (MMA) \cite{emma2023, mma2019}. One of the drawbacks of the MA is that read/write policy module is tightly embedded within the decoder layers. If there are $n$ decoder layers there will be $n$ policy modules. Whereas only the last policy module is used for the read/write decision.

All these monotonic attention based approaches \cite{emma2023, mma2019, milk2019, mocha2018, raffel2017online} try to learn monotonic alignment directly from the data during training in an unsupervised manner. For all possible alignments between source and target tokens, the model optimizes for the alignment that minimizes the token prediction loss and some auxiliary losses like latency loss etc.
For the case of MMA with MoChA and MILK, the training is expensive in terms of compute and memory compared to their non-streaming version. A numerically stable and improved version of MMA called  Efficient Monotonic Multihead Attention (EMMA) was recently introduced in \cite{emma2023} which improves the training and inference strategies, including fine-tuning from an offline translation model and reduction of monotonic alignment variance. However, it still requires more compute resource than the training of a non-streaming model. In our experiment, we could only train the EMMA model on small batch-size due to this issue.

\begin{figure*}[t]
  \includegraphics[width=\linewidth, scale=0.5]{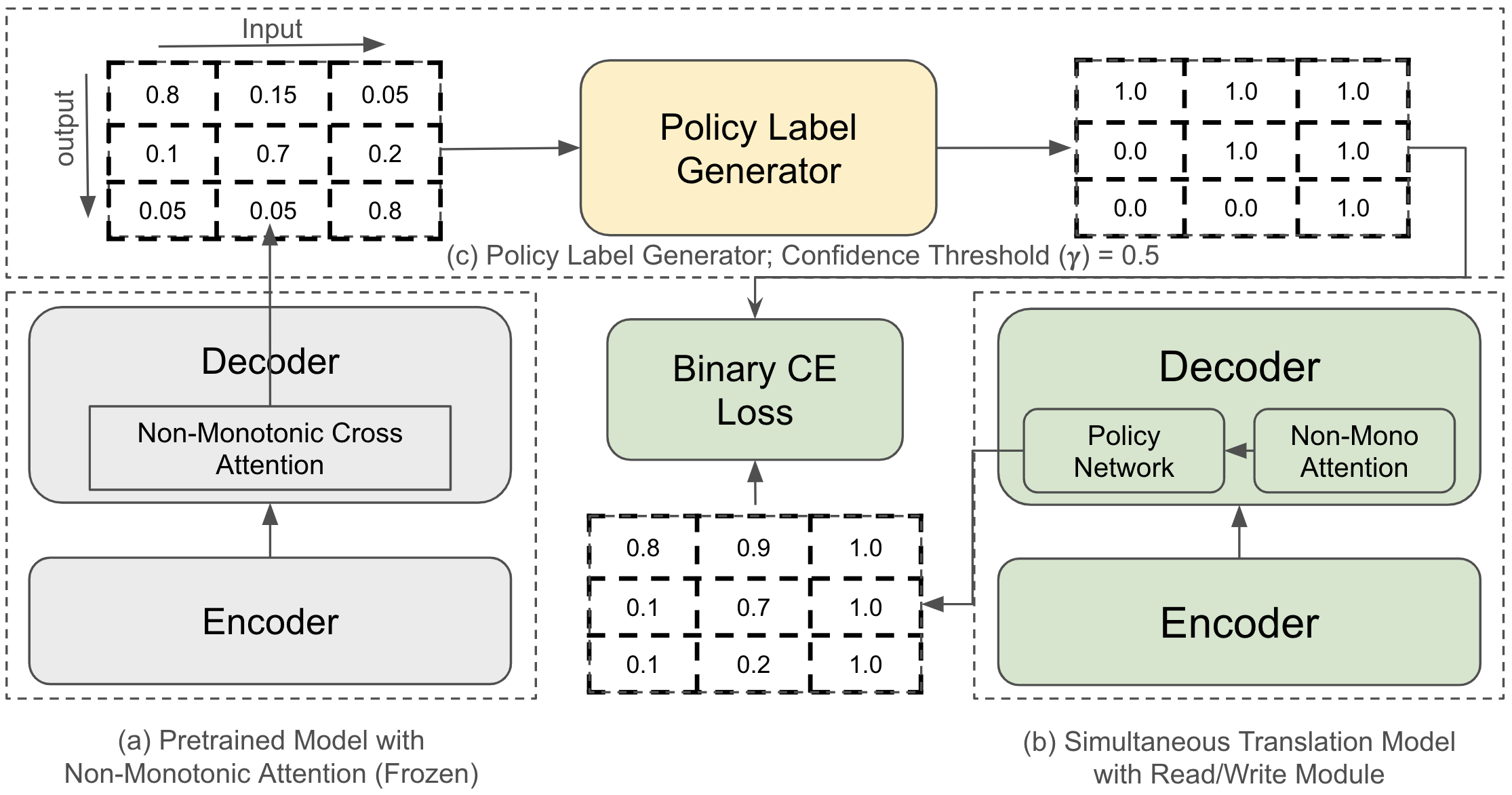} 
  \caption {Read/Write Policy learning using a pretrained non-streaming encoder-decoder model. (a) A pretrained non-streaming model used for generating pseudo labels for training the policy network (this model is not trained). (b) Streaming model with Monotonic Attention and Policy Network. (c) The policy network is learned in a supervised manner using the read/write probability scores generated by Policy Label Generator.}
  \label{architecture}
\end{figure*}

\section{Method}

We enable streaming translation ability in a pretrained non-streaming encoder/decoder model using an additional read/write policy module. Given that the quality of the non-streaming model is currently the upper bound for the streaming model, we propose to learn the read/write policy using the alignment computed from the non-monotonic attention in a supervised manner. We get these supervised labels called pseudo-labels automatically from the non-monotonic attention weights of the pretrained model. The overall architecture of our ALIgnment BAsed STReaming Machine Translation (AliBaStr-MT) is shown in Figure \ref{architecture}. The training of AliBaStr-MT follows the steps
\begin{enumerate}[itemsep=0.5mm]
    \item Train the non-streaming model using cross-entropy objective on output tokens.
    \item Generate the alignment between source and target for the training dataset
    \item Use the alignment to compute pseudo-labels for read/write policy module.
    \item Freeze weights of the non-streaming model, and train the read/write policy module using pseudo-labels.
\end{enumerate}

\subsection{Non-Streaming Full Sentence Translation Model}
The non-streaming model is trained on full sentence context using the following formulation \cite{vaswani2017attention}.  Formally, a model based on standard encoder-decoder architecture transforms an input $x = \{x_1,..,x_{|x|}\}$ into an output sequence $y=\{y_1,...,y_{|y|}\}$ using the following
\begin{align}
    h_j &= Encoder(x_i) \label{enc_eq} \\
    s_i &= Decoder(y_{i-1}, s_{i-1}, c_i) \label{dec_eq}  \\
    y_i &= Output(s_i, c_i) \label{enc_dec_out_eq}
\end{align}
The context vector $c_i$ is computed using a soft attention as follows.
\begin{align}
    e_{i,j} &= Energy(s_{i-1}, h_j) \label{energy_eq} \\
    \alpha_{i,j} &= \frac{exp(e_{i,j})}{\sum_{k=1}^T exp(e_{i,k})} \label{alpha_cxt_vec_eq}  \\
    c_i &= \sum_{j=1}^{|x|}  \alpha_{i,j} h_j \label{cxt_vec_eq}
\end{align}
The attention component of the decoder (termed as non-monotonic attention) in this architecture attends to all the encoder states $h_j : 1 \leq j \leq |x|$  for every decoding time step $ 1 \leq i \leq |y|$. The model is trained by minimizing the negative log likelihood between predicted and reference translation tokens from the training set. 
\subsection{Monotonic Attention and Policy Network}
\cite{emma2023, mma2019, milk2019, mocha2018, raffel2017online} proposed different types of monotonic attention. In terms of architecture, the monotonic attention used in AliBaStr-MT is similar to the non-monotonic attention except that it is equipped with a read/write module which indicates to the model either read further input or attend to the available encoder states and generate translation.  In monotonic attention, the read/write module models the Bernoulli distribution as follows
\begin{align}
     e_{i,j} &= MonotonicEnergy(s_{i-1}, h_j) \label{mono_energy_eq} \\
     p_{i,j} &= Sigmoid(e_{i,j}) \label{mono_sigmoid_eq} \\
     z_{i,j} &\approx Bernoulli(p_{i,j}) 
\end{align}
If $z_{i,j} = 0$, the model reads the input and updates the encoder state, $j$ is incremented while $i$ and decoder states remain unchanged; if $z_{i,j} = 1$, the model writes the translation, updates the decoder states and increments the decoder step $i$ in addition to updating encoder states and encoder step $j$.  The training of monotonic attention involves the following
\begin{itemize}[itemsep=0.5mm]
    \item Generate the $p_{i,j}$ for all $1 \leq i \leq |y|$ and $1 \leq j \leq |x|$. 
    \item Compute the expected alignment using the $p_{i,j}$ ($\alpha_{i,j}$ and $\beta_{i,j}$ as presented in \cite{milk2019}).
    \item  Compute the context vector $c_i$.
    \item  Output translation using $c_i$.
\end{itemize}
Here  the  $p_{i,j}$ are computed first followed by the attention and output projection. In our approach presented in this paper, we follow the standard seq-to-seq training (eq. \ref{enc_eq}-\ref{cxt_vec_eq}).  Our proposed model, as shown in Figure \ref{architecture}(b), is also similar to the non-streaming model of Figure \ref{architecture}(a) except that the decoder has a read/write module making the non-monotonic attention behave as the monotonic attention. The attention module only looks at the input tokens consumed so far. It does not have access to the future input tokens (using a mask as described in section \ref{read_write_section}). The policy network is simply a binary classifier over the encoder and decoder states that predicts the probability based on the similarity between the embeddings.

\subsection{Read/Write Policy Learning}
\label{read_write_section}
The architecture of AliBaStr-MT's read/write module is similar to \cite{mma2019, emma2023} except that read/write module is not part of the  decoder layer. The read/write module is added on top of decoder after all the decoder layers. we compute the current decoder state at step $i$ and compute energy function on the current decoder state contrary to eq. \ref{mono_energy_eq}.
\begin{align}
     e^{'}_{i,j} &= MonotonicEnergy(s_{i}, h_j) \label{mono_energy_eq2} \\
     p^{'}_{i,j} &= Sigmoid(e^{'}_{i,j}) \label{mono_sigmoid_eq2} \\
     z^{'}_{i,j} &\approx Bernoulli(p^{'}_{i,j}) 
\end{align}

If $z_{i,j} = 0$, the model stays at the previous decoder state $i-1$  and goes on to read the next input; if  $z_{i,j} = 1$, the model writes the translation and updates the decoder states to $i$. There is a slight overhead of computing the current decoder state in exchange of better accuracy of read/write decision. This requires at maximum ($|x|+|y|-1$) decoder calls vs. $|y|$ during inference (greedy decoding).
\begin{figure*}
\centering
\includegraphics[scale=0.3]{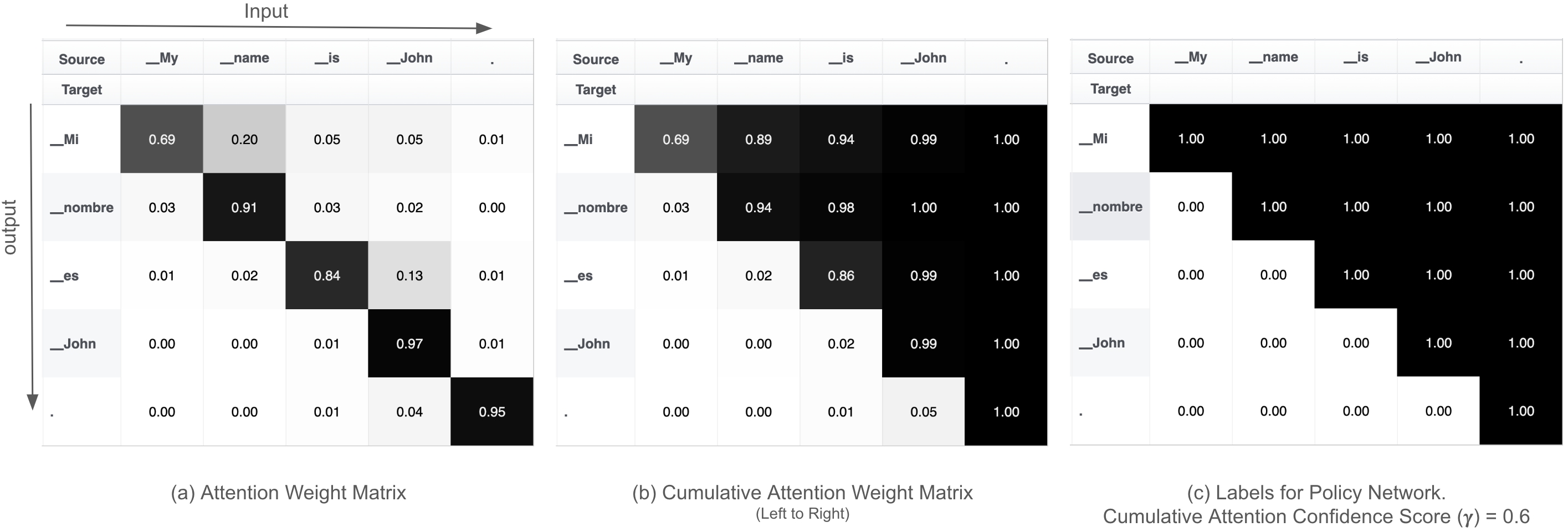} 
\caption{Conversion of Attention weight matrix to Policy label matrix for supervised training of policy network.}
\label{policy_label_matrix_ex_1}
\end{figure*}

For training, the labels are generated by the Policy Label Generator (Figure \ref{architecture} (c)) which uses the attention scores from the pretrained non-streaming model. If we are able to find the point in the attention weight matrix (with some probability threshold < 1) where all the input tokens have been consumed that are required to generate the $i_{th}$ output token, we can construct a label matrix for direct training of the policy network. For example, consider the following input English sentence which is translated in Spanish (using the pretrained non-streaming MT model) as. 
\begin{quote}
My name is John. -> Mi nombre es John. 
\end{quote}
If we consider the attention weight matrix of this sentence as shown in the following Figure \ref{policy_label_matrix_ex_1}(a), we can observe that the soft alignment points for each output token can be converted to policy label matrix using a cumulative attention threshold (Figure \ref{policy_label_matrix_ex_1}(b)). The policy label matrix (Figure \ref{policy_label_matrix_ex_1}(c)) can then be used in the supervised training of the policy network. This example shows a 1:1 mapping of input tokens and output tokens. Figure \ref{policy_label_matrix_ex_2} shows a more complex example.

\subsection{Cumulative Attention Confidence Threshold ($\gamma$)}
\label{cum_attn_conf_thresh}
\begin{figure*}
\centering
\includegraphics[scale=0.28]{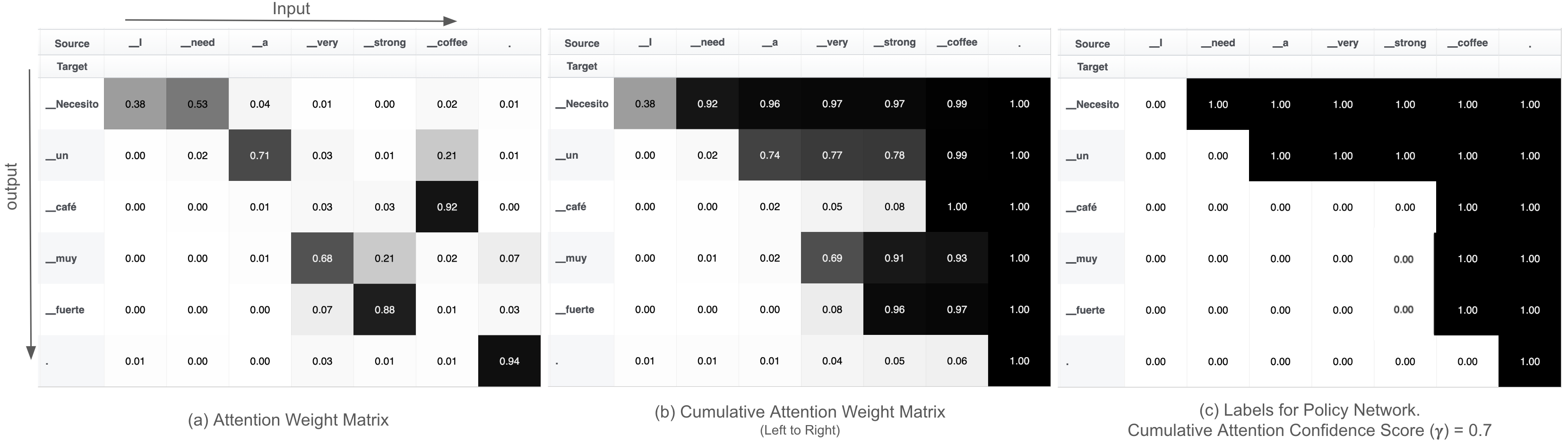} 
\caption{Conversion of Attention weight matrix to Policy label matrix for supervised training of policy network.}
\label{policy_label_matrix_ex_2}
\end{figure*}

To convert the cumulative attention weight matrix into a discreet policy label matrix, we use a attention confidence threshold ($\gamma$). $\gamma$ ranges from 0 to 1. $\gamma$=1 means non-streaming case where policy label generator does not generate "write" label until it has seen full sentence. Any value $\gamma < 1$ enables generation of translation with fewer input tokens. The “1” in the label matrix at position (i, j) indicates a point where the model can generate the $y_{i}$ output token after reading $x_{1..j}$ input tokens.  If $\alpha_{i,j}$ represents the attention weights in the attention matrix $A$, the label matrix $L$ has the following form
\begin{align}
F &= cumsum(A, dim=1) \\
L &= genlabel(F)
\end{align}
where, 
\begin{align}
l_{i,j} &= \begin{cases}
1 & f_{i,j} \geq \gamma \\
0 & Otherwise
\end{cases} \label{eq_l_ij}
\end{align}
Th policy label generator makes sure that policy label matrix is monotonic in terms of input/output i.e. 
\begin{align}
l_{i,j} &= \begin{cases}
0 & l_{i-1,j} = 0 \\
l_{i, j} & Otherwise
\end{cases}
\end{align}
During training, the output $y_{i}$ is not generated with fewer number of input tokens than $y_{i-1}$.  For the example in Figure \ref{policy_label_matrix_ex_1} (b), $\gamma$=0.6 was used to generate the policy label matrix. However, $\gamma$=0.6 is pretty specific to this example. The example in Figure \ref{policy_label_matrix_ex_2} shows $\gamma=0.5$ is a good choice because using this value clearly shows that ”Necesito” cannot be generated until “I need”  is consumed. ”un” can be generated as soon as ”a” is seen. While ”café” will only be generated when ”very strong coffee” has been read. 

Furthermore, there is a possibility that highest $\alpha_{i,j}$ is not covered by the confidence
threshold ($\gamma$) given that $\gamma < 1$. The Policy Label Generator makes sure that the write label is not generated until highest $\alpha_{i,j}$ has been observed. This is achieved by adding a constraint to eq. \ref{eq_l_ij} as follows
\begin{align}
l_{i,j} &= \begin{cases}
1 & f_{i,j} \geq \gamma \:\&\: \underset{j^{'}}{\arg \max}\: (\alpha_{i,j^{'}}) \leq j\\
0 & Otherwise
\end{cases} \label{modified_eq_l_ij}
\end{align}

\subsection{Streaming Beam Search Decoder}
\label{str_beam_search}
Previous papers on simultaneous machine translation (\cite{emma2023, mma2019, milk2019, mocha2018, raffel2017online}) implement only greedy decoding. In our work, we decided implement streaming beam search decoding. This implementation is based on \ref{TODO}.

This implementation presents a couple of challenges such as handling multiple hypotheses with different lengths and model states, deciding when to emit the output, scoring beams, and handling sentence boundaries. The main difference between streaming and non-streaming beam search is that \textbf{each hypothesis in the
beam could have read a different number of tokens and it could have also
written a different number of tokens}. We handle this by processing hypotheses in groups that have seen the same number of tokens from the input. For each group, we continue expanding in a loop hypotheses classified to WRITE until only hypotheses remain that we want to READ. During the expansion step, we score the hypotheses
and always select the top $k$. Once all top $k$ hypotheses want to READ, we
perform that READ simultaneously for all. Once we read all input tokens (reached a sentence boundary), we perform a final WRITE expansion loop until $k$ hypotheses have generated \verb+<EOS>+ token or the maximum output length was reached. Again, at any point in time, all hypotheses in the beam have READ the same input tokens.


Based on our experiments, beam search with beam size three gives 1 BLEU point improvement compared to greedy decoding. This decoder is implemented in a way that we can easily plug in any streaming translation model that returns read/write probability. All of the benchmarks were performed using this decoder.

\section{Evaluation}
We evaluate the performance of our approach on speech-to-text translation task using the cascade of ASR and MT models on English-Spanish language pair. The language pair offers more simultaneous translation opportunity than any other language pair.  We evaluate the models' quality/runtime and present the results for both directions i.e. English->Spanish and Spanish->English. For evaluation, we choose internal English<->Spanish real-life conversation and publicly available FLEURS \cite{conneau2023fleurs} datasets. For training of the MT models, we use publicly available datasets e.g. NLLB, Opensubtitle, CCMatrix, Wikipedia \cite{schwenk2019ccmatrix, fan2021beyond, lison2016opensubtitles2016, wolk2014building} etc. as well as some internal datasets. 

\subsection{Models}
\begin{table*}
\centering
\footnotesize
\begin{tabular}{*{2}{c|}*{2}{c}|*{1}{c}|*{2}{c}|*{1}{c}}
\multirow{3}{*}{Model} & \multirow{3}{*}{Params.} & \multicolumn{3}{c|}{Real-life Conversation} & \multicolumn{3}{c}{FLEURS} \\ \cline{3-8}
 & & \multicolumn{2}{c|}{BLEU} & \multirow{2}{*}{AL} & \multicolumn{2}{c|}{BLEU} & \multirow{2}{*}{AL} \\ \cline{3-4} \cline{6-7} 
 &  & En-Es & Es-En &  & En-Es & Es-En &  \\ \hline 
 \multirow{2}{*}{Non-Streaming} & $Full$ Encoder & 32.62 & 33.20 & \multirow{2}{*}{21.04} & 20.21 & 24.10 & \multirow{2}{*}{33.06} \\ 
 & $Causal$ Encoder & 32.42 & 32.51 & & 19.40 & 23.23 &  \\ \hline
\multirow{3}{*}{Wait-$k$} & $k$=4 & 27.67 & 30.62 & 7.75 & 17.91 & 18.63 & 9.96 \\
 & $k$=6 & 30.13 & 30.90 & 8.59 & 18.65 & 20.1 & 11.26 \\
 & $k$=8 & 30.40 & 31.21 & 9.26 & 19.31 & 21.51 & 12.30 \\ \hline
\multirow{3}{*}{EMMA} & $\lambda$=0.25 & 32.30 & 30.69 & 10.13 & 17.21 & 20.67 & 15.60 \\
 & $\lambda$=0.50 & 30.67 & 31.00 & 9.96 & 17.76 & 20.84 & 15.51 \\
 & $\lambda$=0.75 & 29.42 & 29.58 & 8.96 & 16.70 & 19.75 & 13.30 \\ \hline
\multirow{3}{*}{AliBaStr-MT} & $\delta$=0.50 & 28.55 & 32.09 & 7.58 & 17.56 & 20.87 & 9.94 \\ 
 & $\delta$=0.80 & 29.93 & 32.34 & 8.10 & 18.70 & 22.63 & 11.10 \\ 
 & $\delta$=0.90 & 30.44 & 32.43 & 8.56 & 19.12 & 22.77 & 12.19 \\
\end{tabular}
\caption{BLEU and Average Lag (AL) for Wait-$k$, EMMA and AliBaStr models together with non-streaming baselines. The $k$ is training time hyper-parameter in Wait-$k$, $\lambda$ is a latency weight; a hyper-parameter of EMMA and $\delta$ is the inference time parameter in AliBaStr-MT.}
\label{results_table}
\end{table*}
We train the MT models that are bilingual i.e. that can translate into one of the two target languages; English->Spanish and Spanish-English. A target language $identity$ token is appended as a prefix to input to request desired language at the output. Usually, MT models are train on the written text which is regularly formatted with punctuation, capitalization, and numerals. However, ASR output is normally not formatted where punctuations are missing, everything is lowercased, and numerals are verbalized. To better handle speech-to-text translation scenario, we perform augmentation on the source side. The fraction of the training set is normalized on the source side into ASR like output using a text normalizer. For speech-to-text translation task, we process the raw audio into source text using an internal ASR model.

For better comparison, we choose various state-of-the-art simultaneous/non-simultaneous translation baselines.
\begin{itemize}[itemsep=0.5mm]
    \item \textbf{Non-Streaming Full Encoder}: Encoder/Decoder model with non-monotonic encoder (bidirectional) and non-monotonic decoder having 102M parameters.
    \item \textbf{Non-Streaming Causal Encoder}: Encoder/Decoder model with monotonic encoder (unidirectional) and non-monotonic decoder having 102M parameters.
    \item \textbf{Wait-k}: Encoder/Decoder model with monotonic encoder and decoder with wait-$k$ policy\cite{ma2019stacl}. The model has around 115M parameters.
    \item \textbf{EMMA}: Efficient implementation of MMA approach \cite{emma2023, mma2019} having 124M parameters.
\end{itemize}
All of the models have the same configuration except for the architecture specific differences like presence of tiny read/write module etc. Encoder is based on transformers architecture and has 12-layers. Decoder is also transformer based and has 5-layers. The AliBaStr-MT model has around 103M parameters. The model is similar to Non-Streaming Causal Encoder model except that it has a small read/write module. All the models were trained on 32 A100 machines. The non-streaming models and AliBaStr-MT model were trained with 20K tokens or 20K examples in a batch (whichever is maximum). The wait-$k$ and EMMA could only be trained with (1/10)x of this size. We use Adam optimizer \cite{kingma2014adam} with warm-up and inverse $sqrt$ learning rate schedule. For AliBaStr-MT, we set $\gamma=0.5$ for policy label generation which gave us ample opportunity to tune $\delta$ during inference. For evaluation, we use our streaming beam-search implementation as explained in section \ref{str_beam_search}. The non-streaming models use the standard beam-search. All the models use beam-size of $3$ during evaluation. The input sequence is split into words based on spaces and fed to the streaming decoder one by one. We measure the quality of the model using BLEU and latency of the streaming models using Average Lag (AL)\cite{ma2019stacl} 
which measures the average rate by
which the MT system lags behind an ideal system. 

\subsection{Results}
\begin{figure*}
\centering
\includegraphics[scale=0.38]{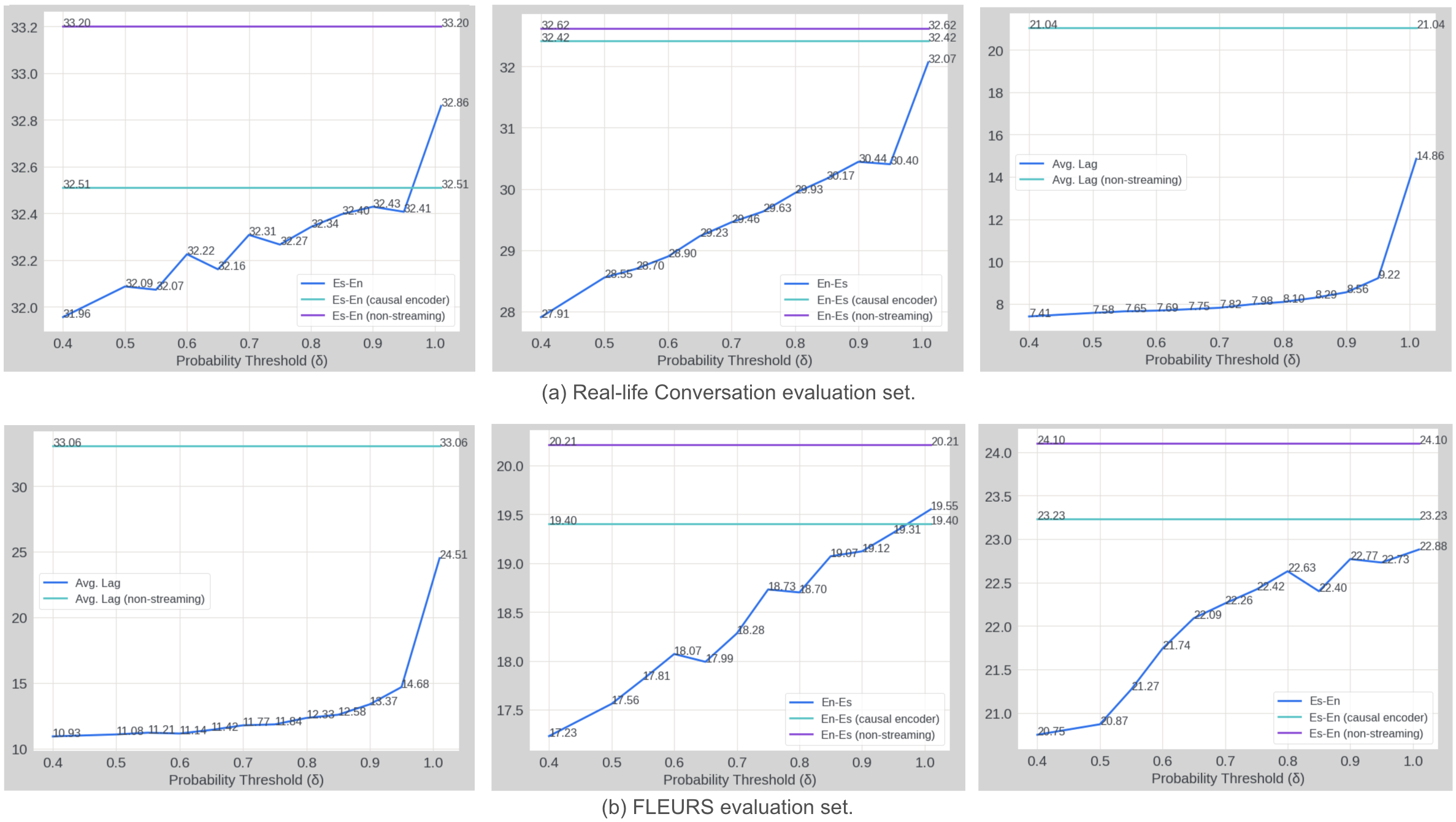} 
\caption{BLEU (En-Es, Es-En) and Average Lag trend against the Read/Write module calibration threshold ($\delta$) on Real-life Conversation and FLEURS evaluation sets.}
\label{bleu_al_trend_vs_cali}
\end{figure*}

The results are presented in the Table \ref{results_table}. The non-streaming full encoder-decoder model is presented as one of the reference baselines. The BLEU and AL of this model is the upper bound for all other models. The non-streaming encoder-decoder model with causal encoder is also a reference baseline which is used as a seed model for the training of the rest of the streaming models i.e. we initialize the common parameters of all the streaming models from this model. None of the streaming models are expected to outperform these two baseline models on BLEU metric. The AL of the streaming models cannot be worse than the AL of the non-streaming models. The AL of the non-streaming model is simply an average number of tokens in the sentence for the evalset. For computing AL, we consider sentences in the evaluation set which are at least 8 tokens in length to avoid results convolution by the shorter sentences specially for the wait-$8$ model. We present the average AL of both translation directions in the table while presenting BLEU separately for each direction.

The Table \ref{results_table} presents wait-$k$ results for three different values of $k=(4, 6, 8)$. The $k$ in wait-$k$ is a training time parameter, therefore each row for wait-$k$ in the table corresponds to entirely different model. For EMMA, the $\lambda$ denotes the latency-weight which also a hyper-parameter and is fixed during training time. The table presents the results for three different models corresponding to $\lambda=(0.25, 0.5, 0.75)$. For AliBaStr-MT, $\delta$ is the inference time parameter which is basically the calibration threshold for the read/write binary classifier. This parameter is tuned during inference. Therefore, the three rows in Table \ref{results_table} corresponding to the AliBaStr-MT, refers to the same model with $\delta=(0.50, 0.80, 0.90)$. As the $\delta$ increases the model delays the translation, hence, BLEU gets better at the cost of increased AL. Our experiments revealed that this property is not inherently supported in wait-$k$ and EMMA models.

The wait-$k$ model performs better in terms of BLEU as the value of $k$ increases while the model latency (AL) gets worse. On the other hand, the $\lambda$ in EMMA is used to control the latency loss component of the total loss. The latency loss was introduced due to EMMA's read/write policy module. The ideal policy that the EMMA's read/write policy module tends to learn is to wait for full input sentence. The latency loss forces the module not to learn the ideal policy. This is also evident by the results for EMMA in the Table \ref{results_table}. As the $\lambda$ gets higher BLEU gets worse while AL improves.

According to the Table \ref{results_table}, the AliBaStr-MT model performance is better than the wait-$k$ and EMMA models. It has better AL than the other two models. The quality of the AliBaStr-MT model (BLEU) is also better than the other two streaming model and has small gap with non-streaming causal encoder model (except for the Real-life Conversation dataset where gap is close to 2 BLEU points on En-Es). The Figure \ref{bleu_al_trend_vs_cali} further highlights effect of quality/latency vs. $\delta$ on the evaluation sets.


\section{Conclusion}

We presented an ALIgnment BAsed STReaming Machine Translation (AliBaStr-MT) technique for training a  simultaneous machine translation model. AliBaStr-MT leverages the alignment information from an encoder-decoder model trained on the full source/target sentence. The alignment information is used to generate the pseudo-labels and the decoder mask for training the read/write module of the AliBaStr-MT model. The experimental evaluation has shown that the model is better than the baseline streaming system in terms of quality (BLEU), latency (Average Lag) and compute needed for training the models. AliBaStr-MT offers the flexibility to control the quality/latency trade-off by tuning the calibration threshold of the read/write module at inference time. AliBaStr-MT is also much more efficient during inference as the read/write module is not coupled with the decoder layer as compared to \cite{milk2019, mma2019, emma2023} where read/write modules are proportional to number of decoder layers.

\section{Limitations}
The known limitations of the ALIgnment BAsed STReaming Machine Translation (AliBaStr-MT) is the need of alignment information which introduces the cumulative attention confidence threshold ($\gamma$) as a hyper-parameter to obtain the decision boundaries for the read/write module. However, during the experiment, we observed that a good value for $\gamma$ is $0.5$ which hardly needs fine-tuning due to the heuristics explained in the section \ref{cum_attn_conf_thresh}.


Furthermore, we tried to fine-tune the read/write module together with the decoder but did not observe much gain. In this setup, we also tried to use the alignment information from the same model which is being fine-tuned (i.e. the model in Figure \ref{architecture}(a) and Figure \ref{architecture}(b) share the parameters) the model starts to hallucinate due to ever-changing pseudo-labels for the training of the policy matrix.





\bibliography{acl_latex}

\appendix



\end{document}